\documentclass{llncs}
\usepackage{pgf}
\usepackage[utf8]{inputenc}
\usepackage[noend]{algpseudocode}
\usepackage{algorithm}
\algrenewcommand\textproc{\texttt}

\usepackage{url}
\usepackage{graphicx,amsmath, amssymb,cite}

\usepackage{multiobjective}
\usepackage[color=blue!29!white,bordercolor=blue!29!white,shadow]{todonotes}

\usepackage{multirow}
\newlength{\wid}
\usepackage{nicefrac}
\usepackage{subfig}
\usepackage{color}
\usepackage{colortbl}

\usepackage{makeidx}  
\usepackage{booktabs}
\begin{document}
%
\title{Anomaly Detection with the Voronoi Diagram Evolutionary Algorithm}
\titlerunning{Anomaly Detection with VorEAl}  
%
\author{Luis~Martí\inst{1}$^{,}$\inst{2} \and Arsene~Fansi-Tchango\inst{3} \and Laurent~Navarro\inst{3} \and Marc~Schoenauer\inst{1}}
\authorrunning{Luis Martí et al.} 
%
\tocauthor{Luis Martí, Arsene Fansi-Tchango, Laurent Navarro and Marc Schoenauer}
\institute{
TAO team, CNRS/INRIA/LRI, Universit\'e Paris-Saclay, Paris, France.
\and
Universidade Federal Flumnense, Niterói (RJ) Brazil.
\and
Thalés Research, Paris, France.
}

\maketitle

\begin{abstract}
This paper presents the Voronoi diagram-based evolutionary algorithm (VorEAl). VorEAl partitions input space in abnormal/normal subsets using Voronoi diagrams. Diagrams are evolved using a multi-objective bio-inspired approach in order to conjointly optimize classification metrics while also being able to represent areas of the data space that are not present in the training dataset. As part of the paper VorEAl is experimentally validated and contrasted with similar approaches.
\end{abstract}
\section{Introduction}

Anomalous Internet traffic detection is a major question of computer network security. Intrusion detection systems (IDSs) \cite{northcutt2002network} have proposed with the intention of tackling this issue. They are meant to protect a network by providing a line of defense that is able to detect and react to network attacks. Two main approaches are used when building an IDS: i) misuse-based and ii) anomaly-based detection. While the former focuses on detecting attacks that follow a known pattern or signature, the latter is interested in building a model representing the system's normal behavior while assuming all deviated activities to be anomalous or intrusions. Because of that fact anomaly detection has received increasing attention in the recent past.

Anomaly detection has been addressed with different approaches (see \cite{chandola2009anomaly} for a survey). Among nature-inspired approaches artificial immune systems (AISs) \cite{Kim2007} have received an special attention.

This paper proposes the Voronoi diagram-based evolutionary algorithm (VorEAl). VorEAl is inspired on AISs and the representations that had been proposed for evolutionary shape design consolidating previous progresses made in this direction \cite{marti-2016}. Its main distinctive feature is that it evolves Voronoi diagram-based representations for normal/abnormal regions of the search space. Such representation offers a flexible and compact alternative to some common representations used in AIS such as hyper-spheres and hyper-rectangles.
%
%
VorEAl applies a multi-objective approach that takes into account the detection accuracy and other especially devised volume-based methods that promotes the emergence of solutions that also adequately represent areas of the input space where no normal data has been received and, therefore, should represent anomalies. As in any multi-objective approach, the algorithm produces a set of trade-off solutions. VorEAl applies a committee approach that is based on the best (in term of \emph{a priori} given set of preferences) subset of those solution.


The paper is organized as follows. Section \ref{sec:foundations} presents the context of AIS and some existing approaches to anomaly detection. Section \ref{sec:VorEAl} introduces the Voronoi representation for abnormal and normal input subsets, together with the variation operators and objective functions used to evolve it and VorEAl as a whole. Section \ref{sec:experiments} introduces our methodology for the experimental validation of VorEAl, also presenting the results of the study and comparing them with other approaches from the literature. Finally, Section \ref{sec:final} discusses the results and sketches some further research directions.

\section{Foundations}\label{sec:foundations}

There has been a consistent interest by the community on proposing nature-inspired approaches to anomaly detection. In this context, AISs have attracted attention as they embody an analogy to the biological immune system. They are particularly appealing for anomaly detection problems as they capture the ability of the biological system of telling apart normal body cells from pathogens. That is, from a computational perspective, they create a model that is able to discriminate between normal (self) and abnormal (non-self) objects. This feature make AISs specially suited to be applied in the context of anomaly-based IDSs.

In order to extend AISs' performance it is necessary to apply algorithms that combine a powerful representation capacity as well as the possibility of adequately adapting that capacity to meet the problem characteristics.

Voronoi diagrams are geometrical constructs that were known by ancient Greeks. Any set of points, known as {\em Voronoi sites}, in a given $n$-dimensional Euclidean space $\cal E$ defines a {\em Voronoi diagram}, i.e., a partition of that space into {\em Voronoi cells}: the cell corresponding to a given site $S$ is the set of points whose closest site is $S$. The boundaries between Voronoi cells are the medians of the $[S_i S_j]$ segments, for neighbor Voronoi sites $S_i$ and $S_j$. Though originally defined in two or three dimensions, there exist several algorithmic procedures to efficiently compute Voronoi diagrams in any dimension. 

Voronoi diagrams offer a compact representation for shapes (surfaces in 2D and volumes in 3D, for instance), by attaching to each Voronoi cell (or, equivalently, to the corresponding Voronoi site), a Boolean label. The resulting Voronoi diagram is a partition of the space into 2 subsets: the ``true'' cells are the shape/volume, and the ``false'' cells are the outside of the shape/volume. The {\em genotype} is here a (variable length) list of labeled Voronoi sites, and the phenotype is the corresponding partition in the space into two subsets. More generally, any piece-wise constant function on the underlying space can be represented by a similar representation by using real-valued labels. Such representation has been successfully used in the context of Evolutionary Optimum Design \cite{schoenauerEurogen95,hamdaVoronoi2001}. In particular, it has been demonstrated that the local complexity of the representation can also be adjusted by evolution: in regions of the space where the shape has a complex boundary, several Voronoi sites will be used, whereas only a few of them will be necessary elsewhere.

In the context of classification, the target phenotypes are partitions of the parameter space into positive and negative examples (in the case of 2 classes), and can hence also be represented by Voronoi diagrams with Boolean labels ---or with labels taken from a finite alphabet in the case of more than 2 classes.

\section{The Voronoi diagrams-based evolutionary algorithm}\label{sec:VorEAl}

We now discuss the building blocks of VorEAl. In particular, we present variation operators, the possible strategies used for evaluating the individuals and how these elements are assembled together to form the algorithm.

\subsection{Variation operators}

The genotypes of Voronoi representations is a variable length list of Voronoi sites $(S_1, \ldots, S_p)$, with $p\in[P_\text{min}, P_\text{max}]$, where each site is defined by its $n$ coordinates in ${\cal E}$. Each site $S$ has an associated label $S.\ell$ that determines how a point that falls within the corresponding cell is classified.

\begin{figure}[t]
\centering
\fbox{
\begin{minipage}{0.74\columnwidth}
\begin{algorithmic}
    \Function{mutate\_voronoi}{$\set{I}, p_\text{s}, p_\text{f}, p_\text{t}, p_+, p_-, \eta$}
    \Statex \quad $\triangleright$ $\set{I}$, individual to be mutated.
    \Statex \quad $\triangleright$ $p_\text{s}\in\left[0,1\right]$, prob. of mutating a site.
    \Statex \quad $\triangleright$ $p_\text{f}\in\left[0,1\right]$, prob. of mutating a site feature (coordinate).
    \Statex \quad $\triangleright$ $p_\text{t}\in\left[0,1\right]$, prob. of changing the label of a site.
    \Statex \quad $\triangleright$ $p_+\in\left[0,1\right]$, prob. of adding a new site.
    \Statex \quad $\triangleright$ $p_-\in\left[0,1\right]$, prob. of removing a site.
    \Statex \quad $\triangleright$ $\eta\in\left(0,\infty\right]$, learning rate.
    \ForAll{$S\in\set{I}$}
      \If{$U[0,1)<p_\text{s}$}
        \ForAll{$x\in S$}
          \If{$U[0,1)<p_\text{f}$}
            \State $x \gets \Call{mutate\_log\_normal}{x, \eta}$
          \EndIf
        \EndFor
        \If{$U[0,1)<p_\text{t}$}
          \State $S.\ell \gets \Call{switch\_label}{S.\ell}$.
        \EndIf
      \EndIf
    \EndFor
    \If{$U[0,1)<p_+$}
      \State $\set{I} \gets \set{I} \cup \left\{\Call{random\_site}{}\right\}$.
    \EndIf
    \If{$U[0,1)<p_-$}
      \State $i \gets U[1,\left|\set{I}\right|)$; $\set{I} \gets \set{I} \setminus \left\{\set{I}(i)\right\}$.
    \EndIf
    \State \Return $\set{I}$, mutated individual.
    \EndFunction
\end{algorithmic}
\end{minipage}}
\caption{Mutation of a Voronoi diagram.}
\label{fig:alg-mutation}
\end{figure}

\paragraph{Mutation Operator}

Several mutation operators can be designed for such a variable-length representation.
\begin{itemize}
\item At the individual level, a Voronoi site can be added, at a randomly chosen position, with a random label; or a randomly chosen Voronoi site can be removed.
\item At the site level, Voronoi sites can be moved around in the space -- and the well-known self-adaptive Gaussian mutation has been chosen here, inspired by Evolution Strategies (see (\ref{SAES}) below); or the label of a Voronoi site can be changed.
\end{itemize}

In the self-adaptive Gaussian mutation \cite{Schwefel}, each coordinate $x$ of each Voronoi site also ``carries'' its own variance $\sigma$ that is used for its Gaussian mutation. Coordinate $x$ undergoes Gaussian mutation with variance $\sigma$ while $\sigma$ undergoes a log-normal mutation with learning rate $\eta$ as follows:
\begin{equation}\label{SAES}
x \leftarrow \sigma {\cal N}(x, 1) \mbox{ and }
\sigma \leftarrow \sigma e^{\eta {\cal N}{0, 1}}
\end{equation}

The different mutation operators are applied according to different probabilities, following the procedure described in Figure \ref{fig:alg-mutation}.

\begin{figure}[t]
  \centering
  \subfloat[Crossover of two Voronoi diagrams, applied with probability $p_c$.\label{fig:alg-cross}]{\fbox{
  \begin{minipage}[b]{0.6\columnwidth}
  \begin{algorithmic}
      \Function{crossover\_voronoi}{$\set{I}_1, \set{I}_2$}
      \Statex \quad $\triangleright$ $\set{I}_1$ and $\set{I}_2$, individuals to mate.
      \Repeat
        \State $\vec{P}\gets\Call{random\_hyperplane}{\set{I}_1\cup\set{I}_2}$.
        \State $\xi^{(1)}_1,\xi^{(1)}_2 \gets \Call{split\_individual}{\set{I}_1, \vec{P}}$.
        \State $\xi^{(2)}_1,\xi^{(2)}_2 \gets \Call{split\_individual}{\set{I}_2, \vec{P}}$.
      \Until{$\xi^{(i)}_k \neq \varnothing, \forall i,k$}
      \State $\set{O}_1 =  \xi^{(1)}_1 \cup \xi^{(2)}_2$; $\set{O}_1.\ell = \set{I}_1.\ell$;
      \State $\set{O}_2 =  \xi^{(2)}_1 \cup \xi^{(1)}_2$; $\set{O}_2.\ell = \set{I}_2.\ell$.
      \State \Return $\set{O}_1,\set{O}_2$, offspring.
      \EndFunction
  \end{algorithmic}
  \end{minipage}}}\hfill
  \subfloat[Example of crossover of two Voronoi genotype individuals in 2D.\label{fig:mating}]{\includegraphics[width=0.35\columnwidth]{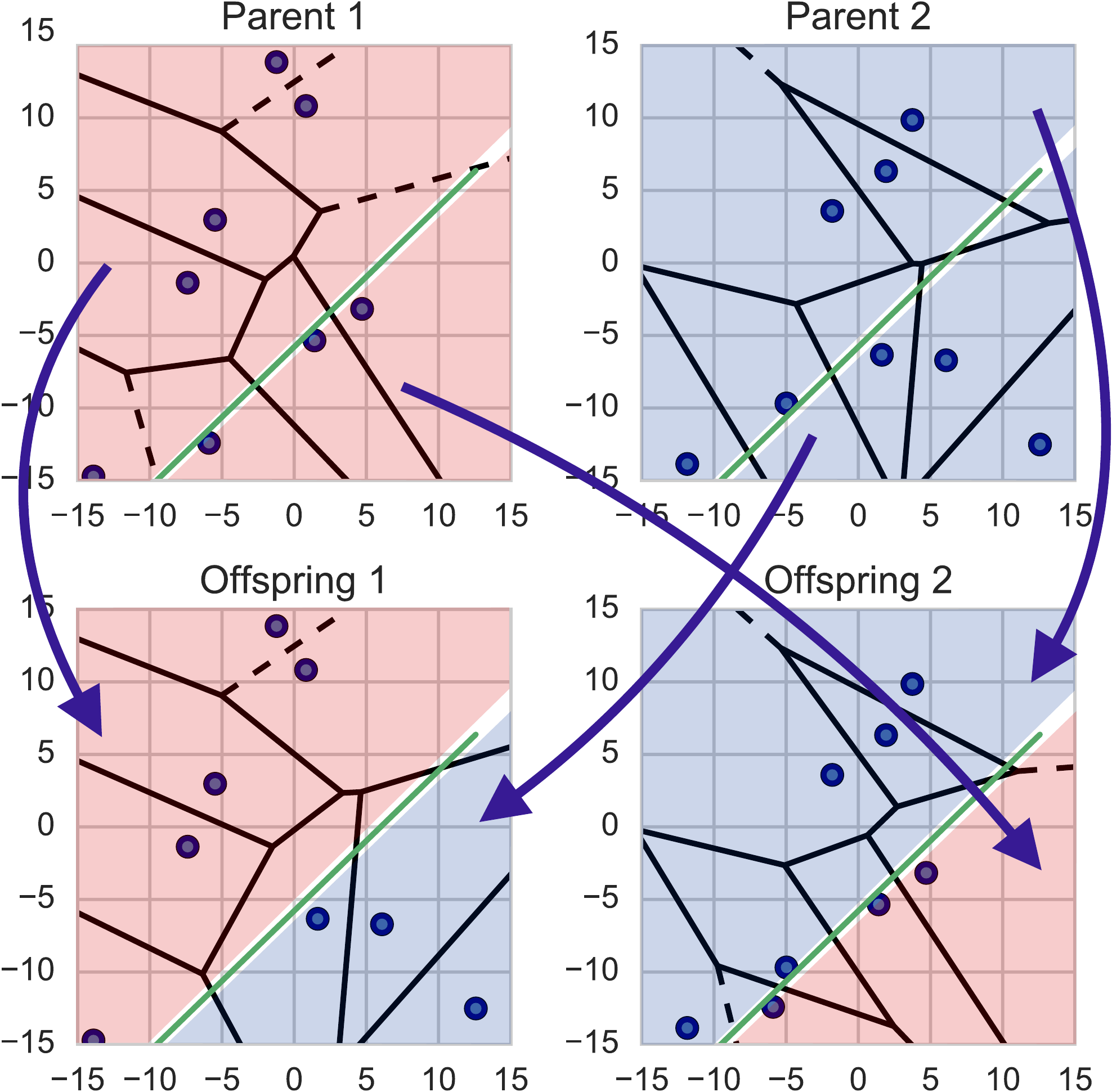}}
  \caption{Crossover operator for Voronoi diagrams.}
\end{figure}

\paragraph{Crossover Operator} The crossover operator for Voronoi representation should not simply exchange some Voronoi sites between both parents, but should respect the locality of the representation. Voronoi sites that are close to each other should have more chance to stay together than Voronoi sites that are far apart. This is achieved by the geometric crossover that operates on two (randomly selected) parents by creating a random cutting hyperplane, and exchanges the Voronoi sites from both sides of the hyperplane. The Voronoi diagrams are of course reconstructed after the crossover. This procedure is described in detail in Figure \ref{fig:alg-cross}. A two-dimensional example is given in Figure \ref{fig:mating}.

\subsection{Objectives and fitness assignment}\label{sec:objectives}

Anomaly detection can be posed as a particular case of classification problem where data items must be tagged either as ``normal'' or ``anomalous''. That relying on a dataset $\Psi = \left\{\vec{x}^{(i)}, y^{(i)}\right\}$ where, without loss of generality we can state that $\vec{x}\in\mathbb{R}^n$ and $y^{(i)}\in\left\{\mathrm{normal};\mathrm{anomaly}\right\}$ obtain a classifier that correctly detects instances that correspond to each of the two categories. Because of this fact the existing metrics devised to assess the quality of a classification algorithm are also applicable in this context. For this particular problem, the most relevant metrics are accuracy, recall and specificity, although many more could also be of use. 
Accuracy seems the best choice in the general case, as one wants to correctly identify all examples. But when dealing with anomalies, the dataset is generally highly imbalanced, as normally there are fewer anomalous instances than `normal' ones. If only the classification accuracy is used, the error contribution of the anomalies will be reduced and hence the model will be biased to not regard them.



Furthermore, as already mentioned, the anomaly detection problem requires that the classifier is not only able to correctly classify the ``normal'' and ``anomalous'' instances present in the training dataset but is also capable of detecting when a given input falls in an area that was not covered by data of the training set and, therefore, also can be interpreted as an anomaly.

It is possible to prompt the Voronoi diagrams (individuals) to represent the known data in a form as compact as possible by expressing that as the relation between the volumes of the Voronoi cell and the convex hull of the training data that it contains. Let $\mathcal{I}=\left\{S_i, i=1\dots n_{\cal I}\right\}$ be a Voronoi diagram, and, for each cell $C_i$, let $v_i\in\mathbb{R}$ be its volume and $\set{D}_{i}$ the set of data points it contains, i.e., $\set{D}_{i}=\left\{\vec{x}\in\Psi; d(x,S_i)\leq d(x, S_j) \forall i\neq j\right\}$, $d$ being the $n$-dimensional Euclidian distance. We can then define the individual compactness as the sum, for each cell, of the ratio of the volume of the convex hulls of $\set{D}_i$ and the volume of the cell,
\begin{equation}
\mathrm{C}(\set{I})=
\left\{
\begin{array}{cl}
\sum_{i}\frac{\text{volume}\left(\text{convex\_hull}\left(\set{D}_i\right)\right)}{v_i} & \text{if } \left|\set{D}_i\right|>n,\\
0 & \text{in other case}.\\
\end{array}\right.
\end{equation}
It could be hypothesized that the previous formulation can be improved by adding a multiplicative term that counts the number of elements in $\set{D}_i$, resulting in the multiplicative compactness
\begin{equation}
\mathrm{C}_\text{mult}(\set{I})=
\left\{
\begin{array}{cl}
\sum_{i}\left(\left|\set{D}_i\right| -n\right)\frac{\text{volume}\left(\text{convex\_hull}\left(\set{D}_i\right)\right)}{v_i} & \text{if } \left|\set{D}_i\right|>n,\\
0 & \text{in other case}.\\
\end{array}\right.
\end{equation}
In both cases, maximizing the compactness will produce cells that contain the data in a form as tight as possible. Those compactness objectives can be complemented by one that promotes the existence of empty cells that represent areas of the input domain that are now present in the training data. Such objective would take care of sites with small $\set{D}_i$'s and promote that they become empty as the evolution takes place. A form of representing this is by computing the total volume of cells with an anomaly label of an individual and rate it by the number of elements it contains,
\begin{equation}
  \mathrm{EV}(\set{I})=\sum_{i, S_i.\ell=\text{anomaly}} \frac{v_i}{1+2\ln(\left|\set{D}_i\right|+1)}\,.
\end{equation}
Consequently, it is obvious that it is necessary to jointly address all of those objectives. Therefore, a multi-objective optimization approach will empower the algorithm with the capacity to address all the requirements of the task at the same time.



\subsection{Algorithm description}\label{sec:algo}

VorEAl consolidates the previous components as an algorithm that constructs a classification model. The algorithm starts by creating an initial random population $\set{P}_0$ of $n_\text{pop}$ individuals. At a given iteration $t$, individuals in the population $\set{P}_t$ are then mutated and mated using operators described above and thus producing an offspring population $\set{P}_\text{off}$ that consists of $n_\text{off}$ individuals. At this point, individuals that have not yet been evaluated are presented with the dataset and the values of the different objective functions are calculated. In this work, we compute the accuracy, recall and specificity, but it should be noted that others are available. From the union of $\set{P}_t$ and $\set{P}_\text{off}$, the best $n_\text{pop}$ are selected using the non-dominated sorting selection of NSGA-II \cite{Deb02}.


This process repeats until the stopping criterion of the algorithm is met. When that happens, the algorithm has a final population $\set{P}_\text{final}$ from which the best individual(s) can be selected to represent the `self' of the AIS. This a non-trivial task as it implies taking into account the different conflictive objectives. In this work, we select a committee of individuals $\set{P}_\text{committee}\subseteq\set{P}_\text{final}$ that contains the $\rho$-percent of $\set{P}_\text{final}$ with the highest accuracy. Hence, the classifier returns the most voted decision among the members of $\set{P}_\text{committee}$.

\begin{figure*}[t]
  \centering
  \subfloat[Training datasets.]{\includegraphics[width=0.47\textwidth]{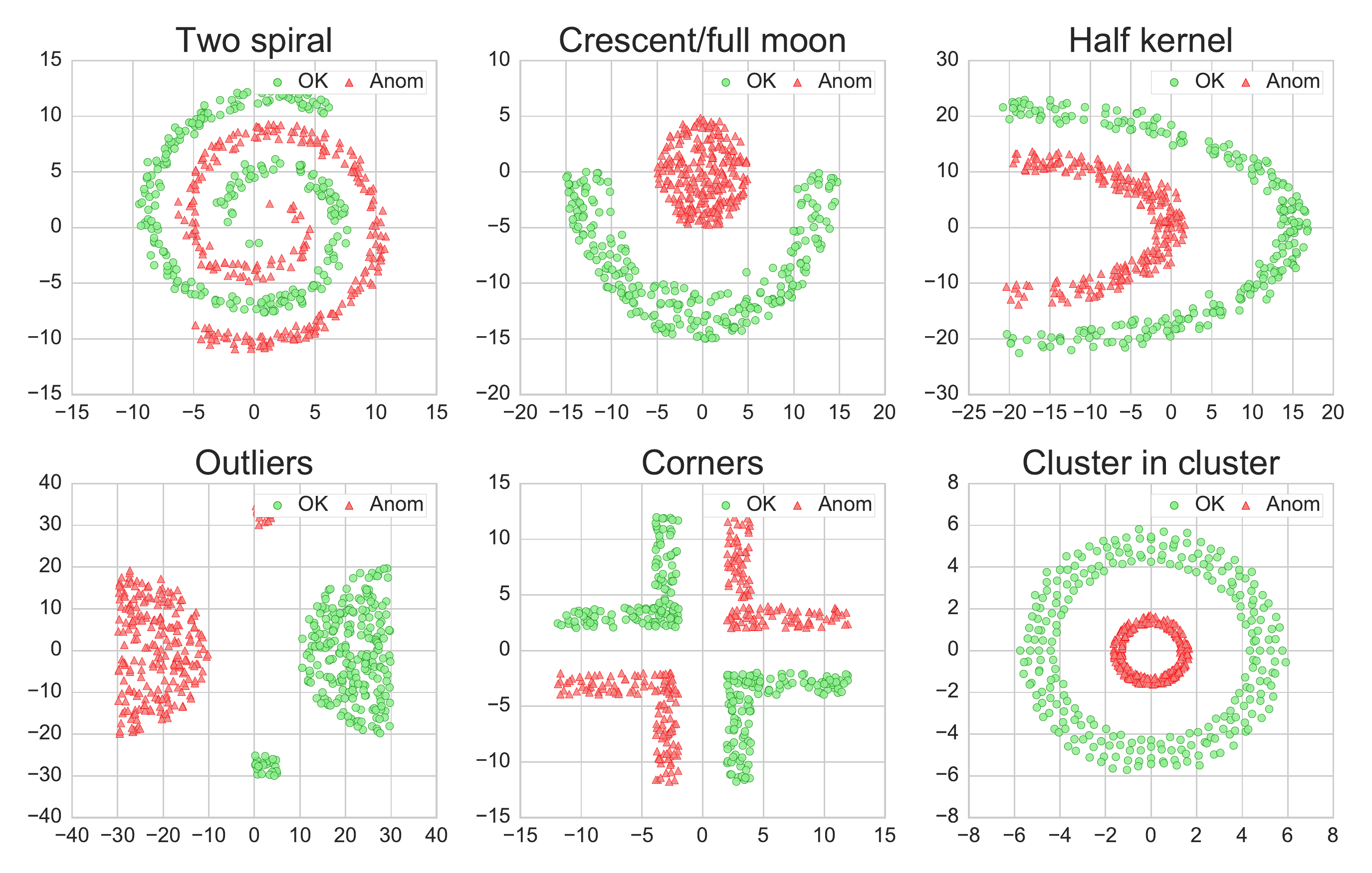}}\hfill
  \subfloat[Test datasets.]{\includegraphics[width=0.47\textwidth]{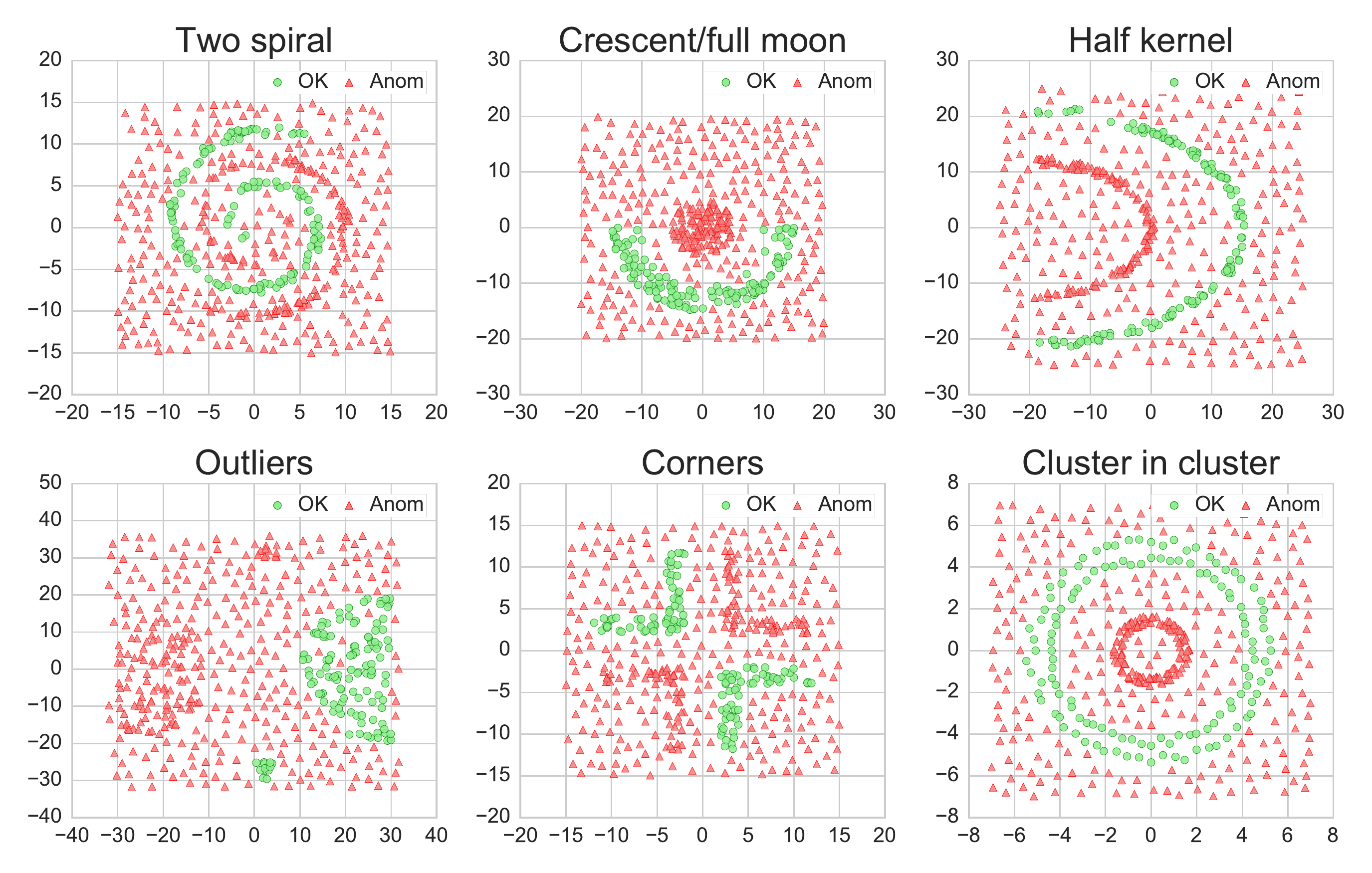}}
  \caption{Training and testing datasets. Test set anomalies present in the test datasets are generated using the procedure described in Section \ref{sec:experiments}.}
  \label{fig:problems}
\end{figure*}

\section{Experimental study}\label{sec:experiments}

The previous discussion and proposal must be complemented by a set of experiments that establish the validity of VorEAl and studies the impacts of the different components presented. That is the focus of this section.

One of the main questions regarding VorEAl is at what point a multi-objective affinity function would actually generate better results at an admissible cost. It could be argued that there exists the possibility that adding more objectives would just make the search process more complex and resource demanding.

An important matter to be clarified was the impact of each of the objectives presented in previous section. For that reason different combinations were tested. In particular, we tested accuracy and compactness (a/c); accuracy, compactness and total empty volume (a/c/t); accuracy and multiplicative compactness (a/m) and accuracy, multiplicative compactness and total empty volume (a/m/t).

In order to provide grounds for comparison with similar approaches as well as well-known approaches, other methods were included in the experiments. In particular, we included the negative selection algorithm (NSA) \cite{Ji2004a} using both variable-sized hyper-spheres and hyper-rectangles. For fair comparisons, we applied the NSA${}^{+}_{sp}$ and NSA${}^{+}_{re}$  in which non-self training samples are subsequently used to enrich the detector library generated by NSA.

Similarly, we have included in the experiments two well-known classifiers: one-class vector machines (SVMs) \cite{tax2004support} and the naïve Bayes classifier.

The experiments involved six classification benchmarks problems: the `two spiral', `crescent and full moon', `half densities', `corners', `outliers' and `cluster in cluster' problems. They have the advantage that they can be visualized in 2D while still posing a substantial challenge to the algorithm. One key element that must be addressed is the ability of the method to detect anomalies that were present in the original dataset and also those that were not present. Six tests were prepared with that goal in mind by adding random anomaly data in the areas that did not had any data in the training dataset. The resulting training and test datasets can be observed in Figure \ref{fig:problems}. Besides fixing these parameters we limited the population to 100 individuals and ran the algorithms for 500 generations. The rest of the parameters are tuned using a grid search procedure on a reduced-size problem. The same parameters were used for all problems. The mutation of the parameters were $p_\text{s}=0.5$, $p_\text{f}=0.5$, $p_\text{t}=0.1$, $p_+=0.2$, $p_-=0.1$ and $\eta=0.5$, while the mating probability was 0.5, the minimum and maximum number of sites in an individual was set to 20 and 100, respectively and the committee selection percentile ($\rho$) was set to 5\% of the population.

\begin{figure}[t] 
\centering
\includegraphics[width=\columnwidth]{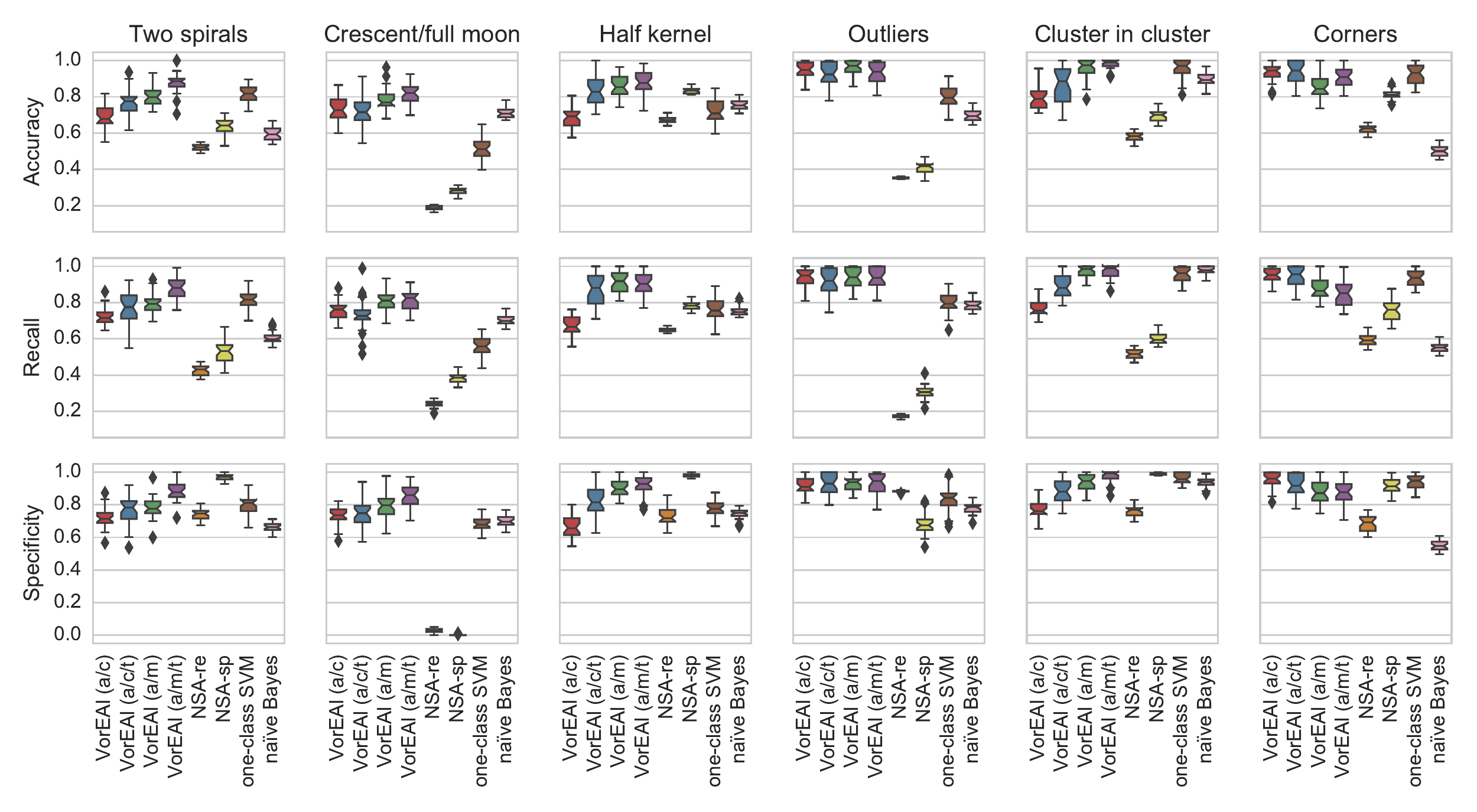}
\caption{Box plots of the experimental evaluations on the anomaly detection test sets.}
\label{fig:boxplots-problems}
\end{figure}

The stochastic nature of the algorithms being analyzed calls for the use of an experimental methodology that relies on statistical hypothesis tests. Using those tests, we are able to determine in a statistical sound way if one algorithm instance outperforms another. The topic of assessing stochastic classification algorithms is studied in depth in \cite{garcia-2008:comparison}. There, it is shown that the Bergmann--Hommel 
procedure is the most suitable for our class of problem.
In all cases, we have used a base level of significance of 0.05 and we run the same experiment instances 50 times. The results of this experiments are shown as box plots in Figure \ref{fig:boxplots-problems}. It can be inferred from those plots that the three-objective form of with accuracy, multiplicative compactness, and total empty volume VorEAl yielded the best results.

\newcommand{\nc}{\cellcolor{red!29}$-$}
\newcommand{\pc}{\cellcolor{green!29}$+$}
\newcommand{\no}{\cellcolor{gray!29}$\cdot$}
\newcommand{\ii}{\cellcolor{blue!29}$\sim$}

\newcolumntype{C}[1]{>{\centering\let\newline\\\arraybackslash\hspace{0pt}}p{#1}}

\begin{table}[t]
  \caption{Summary of the outcome of the statistical hypothesis tests for each problem and performance indicator. When an algorithm in the row has been significantly better than the one in the column the corresponding cell is marked with a ``$+$''. If it has been worst then the cell contains a ``$-$''. Cases where no significant difference was established are identified with a ``$\sim$''.}
  \label{tbl:stats}
  \resizebox{\textwidth}{!}{
    \begin{tabular}{@{}c@{}c@{}c@{}c@{}c@{}c@{}}
      \begin{tabular}{crC{\wid}C{\wid}C{\wid}C{\wid}C{\wid}C{\wid}C{\wid}C{\wid}}
& & \multicolumn{8}{c}{\textbf{Two Spiral}}\\
\cmidrule[\heavyrulewidth](lr){3-10}
& &
\rotatebox{90}{VorEAl (a/c)} &
\rotatebox{90}{VorEAl (a/c/t)} &
\rotatebox{90}{VorEAl (a/m)} &
\rotatebox{90}{VorEAl (a/m/t)} &
\rotatebox{90}{NSA-re} &
\rotatebox{90}{NSA-sp} &
\rotatebox{90}{one-class SVM} &
\rotatebox{90}{naïve Bayes} \\
\midrule
\multirow{8}{*}{\rotatebox{90}{\textbf{Accuracy}}} &
VorEAl (a/c)   &          \no &            \nc &          \nc &            \nc &    \pc &    \pc &           \nc &         \pc \\
& VorEAl (a/c/t) &          \pc &            \no &          \nc &            \nc &    \pc &    \pc &           \nc &         \pc \\
& VorEAl (a/m)   &          \pc &            \pc &          \no &            \nc &    \pc &    \pc &           \ii &         \pc \\
& VorEAl (a/m/t) &          \pc &            \pc &          \pc &            \no &    \pc &    \pc &           \pc &         \pc \\
& NSA-re         &          \nc &            \nc &          \nc &            \nc &    \no &    \nc &           \nc &         \nc \\
& NSA-sp         &          \nc &            \nc &          \nc &            \nc &    \pc &    \no &           \nc &         \pc \\
& one-class SVM  &          \pc &            \pc &          \ii &            \nc &    \pc &    \pc &           \no &         \pc \\
& naïve Bayes    &          \nc &            \nc &          \nc &            \nc &    \pc &    \nc &           \nc &         \no \\
\midrule
\end{tabular}
 &
      \begin{tabular}{C{\wid}C{\wid}C{\wid}C{\wid}C{\wid}C{\wid}C{\wid}C{\wid}}
\multicolumn{8}{c}{\textbf{Crescent/full moon}}\\
\cmidrule[\heavyrulewidth](lr){1-8}
\rotatebox{90}{VorEAl (a/c)} &
\rotatebox{90}{VorEAl (a/c/t)} &
\rotatebox{90}{VorEAl (a/m)} &
\rotatebox{90}{VorEAl (a/m/t)} &
\rotatebox{90}{NSA-re} &
\rotatebox{90}{NSA-sp} &
\rotatebox{90}{one-class SVM} &
\rotatebox{90}{naïve Bayes} \\
\midrule
         \no &            \ii &          \nc &            \nc &    \pc &    \pc &           \pc &         \ii \\
         \ii &            \no &          \nc &            \nc &    \pc &    \pc &           \pc &         \ii \\
         \pc &            \pc &          \no &            \nc &    \pc &    \pc &           \pc &         \pc \\
         \pc &            \pc &          \pc &            \no &    \pc &    \pc &           \pc &         \pc \\
         \nc &            \nc &          \nc &            \nc &    \no &    \nc &           \nc &         \nc \\
         \nc &            \nc &          \nc &            \nc &    \pc &    \no &           \nc &         \nc \\
         \nc &            \nc &          \nc &            \nc &    \pc &    \pc &           \no &         \nc \\
         \ii &            \ii &          \nc &            \nc &    \pc &    \pc &           \pc &         \no \\
\midrule
\end{tabular}
 &
      \begin{tabular}{C{\wid}C{\wid}C{\wid}C{\wid}C{\wid}C{\wid}C{\wid}C{\wid}}
\multicolumn{8}{c}{\textbf{Half kernel}}\\
\cmidrule[\heavyrulewidth](lr){1-8}
\rotatebox{90}{VorEAl (a/c)} &
\rotatebox{90}{VorEAl (a/c/t)} &
\rotatebox{90}{VorEAl (a/m)} &
\rotatebox{90}{VorEAl (a/m/t)} &
\rotatebox{90}{NSA-re} &
\rotatebox{90}{NSA-sp} &
\rotatebox{90}{one-class SVM} &
\rotatebox{90}{naïve Bayes} \\
\midrule
         \no &            \nc &          \nc &            \nc &    \pc &    \nc &           \nc &         \nc \\
         \pc &            \no &          \ii &            \nc &    \pc &    \ii &           \pc &         \pc \\
         \pc &            \ii &          \no &            \ii &    \pc &    \ii &           \pc &         \pc \\
         \pc &            \pc &          \ii &            \no &    \pc &    \pc &           \pc &         \pc \\
         \nc &            \nc &          \nc &            \nc &    \no &    \nc &           \nc &         \nc \\
         \pc &            \ii &          \ii &            \nc &    \pc &    \no &           \pc &         \pc \\
         \pc &            \nc &          \nc &            \nc &    \pc &    \nc &           \no &         \nc \\
         \pc &            \nc &          \nc &            \nc &    \pc &    \nc &           \pc &         \no \\
\midrule
\end{tabular}
 &
      \begin{tabular}{C{\wid}C{\wid}C{\wid}C{\wid}C{\wid}C{\wid}C{\wid}C{\wid}}
\multicolumn{8}{c}{\textbf{Outliers}}\\
\cmidrule[\heavyrulewidth](lr){1-8}
\rotatebox{90}{VorEAl (a/c)} &
\rotatebox{90}{VorEAl (a/c/t)} &
\rotatebox{90}{VorEAl (a/m)} &
\rotatebox{90}{VorEAl (a/m/t)} &
\rotatebox{90}{NSA-re} &
\rotatebox{90}{NSA-sp} &
\rotatebox{90}{one-class SVM} &
\rotatebox{90}{naïve Bayes} \\
\midrule
         \no &            \ii &          \ii &            \ii &    \pc &    \pc &           \pc &         \pc \\
         \ii &            \no &          \nc &            \ii &    \pc &    \pc &           \pc &         \pc \\
         \ii &            \pc &          \no &            \pc &    \pc &    \pc &           \pc &         \pc \\
         \ii &            \ii &          \nc &            \no &    \pc &    \pc &           \pc &         \pc \\
         \nc &            \nc &          \nc &            \nc &    \no &    \nc &           \nc &         \nc \\
         \nc &            \nc &          \nc &            \nc &    \pc &    \no &           \nc &         \nc \\
         \nc &            \nc &          \nc &            \nc &    \pc &    \pc &           \no &         \pc \\
         \nc &            \nc &          \nc &            \nc &    \pc &    \pc &           \nc &         \no \\
\midrule
\end{tabular}
 &
      \begin{tabular}{C{\wid}C{\wid}C{\wid}C{\wid}C{\wid}C{\wid}C{\wid}C{\wid}}
\multicolumn{8}{c}{\textbf{Cluster in cluster}}\\
\cmidrule[\heavyrulewidth](lr){1-8}
\rotatebox{90}{VorEAl (a/c)} &
\rotatebox{90}{VorEAl (a/c/t)} &
\rotatebox{90}{VorEAl (a/m)} &
\rotatebox{90}{VorEAl (a/m/t)} &
\rotatebox{90}{NSA-re} &
\rotatebox{90}{NSA-sp} &
\rotatebox{90}{one-class SVM} &
\rotatebox{90}{naïve Bayes} \\
\midrule
         \no &            \nc &          \nc &            \nc &    \pc &    \pc &           \nc &         \nc \\
         \pc &            \no &          \nc &            \nc &    \pc &    \pc &           \nc &         \ii \\
         \pc &            \pc &          \no &            \nc &    \pc &    \pc &           \ii &         \pc \\
         \pc &            \pc &          \pc &            \no &    \pc &    \pc &           \pc &         \pc \\
         \nc &            \nc &          \nc &            \nc &    \no &    \nc &           \nc &         \nc \\
         \nc &            \nc &          \nc &            \nc &    \pc &    \no &           \nc &         \nc \\
         \pc &            \pc &          \ii &            \nc &    \pc &    \pc &           \no &         \pc \\
         \pc &            \ii &          \nc &            \nc &    \pc &    \pc &           \nc &         \no \\
\midrule
\end{tabular}
 &
      \begin{tabular}{C{\wid}C{\wid}C{\wid}C{\wid}C{\wid}C{\wid}C{\wid}C{\wid}}
\multicolumn{8}{c}{\textbf{Corners}}\\
\midrule
\rotatebox{90}{VorEAl (a/c)} &
\rotatebox{90}{VorEAl (a/c/t)} &
\rotatebox{90}{VorEAl (a/m)} &
\rotatebox{90}{VorEAl (a/m/t)} &
\rotatebox{90}{NSA-re} &
\rotatebox{90}{NSA-sp} &
\rotatebox{90}{one-class SVM} &
\rotatebox{90}{naïve Bayes} \\
\midrule
         \no &            \ii &          \pc &            \pc &    \pc &    \pc &           \ii &         \pc \\
         \ii &            \no &          \pc &            \ii &    \pc &    \pc &           \ii &         \pc \\
         \nc &            \nc &          \no &            \nc &    \pc &    \pc &           \nc &         \pc \\
         \nc &            \ii &          \pc &            \no &    \pc &    \pc &           \ii &         \pc \\
         \nc &            \nc &          \nc &            \nc &    \no &    \nc &           \nc &         \pc \\
         \nc &            \nc &          \nc &            \nc &    \pc &    \no &           \nc &         \pc \\
         \ii &            \ii &          \pc &            \ii &    \pc &    \pc &           \no &         \pc \\
         \nc &            \nc &          \nc &            \nc &    \nc &    \nc &           \nc &         \no \\
\midrule
\end{tabular}
\\
      \begin{tabular}{rrC{\wid}C{\wid}C{\wid}C{\wid}C{\wid}C{\wid}C{\wid}C{\wid}}
\multirow{8}{*}{\rotatebox{90}{\textbf{Recall}}} &
VorEAl (a/c)   &          \no &            \nc &          \nc &            \nc &    \pc &    \pc &           \nc &         \pc \\
& VorEAl (a/c/t) &          \pc &            \no &          \ii &            \nc &    \pc &    \pc &           \nc &         \pc \\
& VorEAl (a/m)   &          \pc &            \ii &          \no &            \nc &    \pc &    \pc &           \ii &         \pc \\
& VorEAl (a/m/t) &          \pc &            \pc &          \pc &            \no &    \pc &    \pc &           \pc &         \pc \\
& NSA-re         &          \nc &            \nc &          \nc &            \nc &    \no &    \nc &           \nc &         \nc \\
& NSA-sp         &          \nc &            \nc &          \nc &            \nc &    \pc &    \no &           \nc &         \nc \\
& one-class SVM  &          \pc &            \pc &          \ii &            \nc &    \pc &    \pc &           \no &         \pc \\
& naïve Bayes    &          \nc &            \nc &          \nc &            \nc &    \pc &    \pc &           \nc &         \no \\
\midrule
\end{tabular}
 &
      \begin{tabular}{C{\wid}C{\wid}C{\wid}C{\wid}C{\wid}C{\wid}C{\wid}C{\wid}}
         \no &            \ii &          \nc &            \nc &    \pc &    \pc &           \pc &         \pc \\
         \ii &            \no &          \nc &            \nc &    \pc &    \pc &           \pc &         \pc \\
         \pc &            \pc &          \no &            \ii &    \pc &    \pc &           \pc &         \pc \\
         \pc &            \pc &          \ii &            \no &    \pc &    \pc &           \pc &         \pc \\
         \nc &            \nc &          \nc &            \nc &    \no &    \nc &           \nc &         \nc \\
         \nc &            \nc &          \nc &            \nc &    \pc &    \no &           \nc &         \nc \\
         \nc &            \nc &          \nc &            \nc &    \pc &    \pc &           \no &         \nc \\
         \nc &            \nc &          \nc &            \nc &    \pc &    \pc &           \pc &         \no \\
\midrule
\end{tabular}
 &
      \begin{tabular}{C{\wid}C{\wid}C{\wid}C{\wid}C{\wid}C{\wid}C{\wid}C{\wid}}
         \no &            \nc &          \nc &            \nc &    \pc &    \nc &           \nc &         \nc \\
         \pc &            \no &          \ii &            \ii &    \pc &    \pc &           \pc &         \pc \\
         \pc &            \ii &          \no &            \ii &    \pc &    \pc &           \pc &         \pc \\
         \pc &            \ii &          \ii &            \no &    \pc &    \pc &           \pc &         \pc \\
         \nc &            \nc &          \nc &            \nc &    \no &    \nc &           \nc &         \nc \\
         \pc &            \nc &          \nc &            \nc &    \pc &    \no &           \pc &         \pc \\
         \pc &            \nc &          \nc &            \nc &    \pc &    \nc &           \no &         \ii \\
         \pc &            \nc &          \nc &            \nc &    \pc &    \nc &           \ii &         \no \\
\midrule
\end{tabular}
 &
      \begin{tabular}{C{\wid}C{\wid}C{\wid}C{\wid}C{\wid}C{\wid}C{\wid}C{\wid}}
         \no &            \ii &          \ii &            \ii &    \pc &    \pc &           \pc &         \pc \\
         \ii &            \no &          \ii &            \ii &    \pc &    \pc &           \pc &         \pc \\
         \ii &            \ii &          \no &            \ii &    \pc &    \pc &           \pc &         \pc \\
         \ii &            \ii &          \ii &            \no &    \pc &    \pc &           \pc &         \pc \\
         \nc &            \nc &          \nc &            \nc &    \no &    \nc &           \nc &         \nc \\
         \nc &            \nc &          \nc &            \nc &    \pc &    \no &           \nc &         \nc \\
         \nc &            \nc &          \nc &            \nc &    \pc &    \pc &           \no &         \ii \\
         \nc &            \nc &          \nc &            \nc &    \pc &    \pc &           \ii &         \no \\
\midrule
\end{tabular}
 &
      \begin{tabular}{C{\wid}C{\wid}C{\wid}C{\wid}C{\wid}C{\wid}C{\wid}C{\wid}}
         \no &            \nc &          \nc &            \nc &    \pc &    \pc &           \nc &         \nc \\
         \pc &            \no &          \nc &            \nc &    \pc &    \pc &           \nc &         \nc \\
         \pc &            \pc &          \no &            \ii &    \pc &    \pc &           \pc &         \ii \\
         \pc &            \pc &          \ii &            \no &    \pc &    \pc &           \pc &         \ii \\
         \nc &            \nc &          \nc &            \nc &    \no &    \nc &           \nc &         \nc \\
         \nc &            \nc &          \nc &            \nc &    \pc &    \no &           \nc &         \nc \\
         \pc &            \pc &          \nc &            \nc &    \pc &    \pc &           \no &         \nc \\
         \pc &            \pc &          \ii &            \ii &    \pc &    \pc &           \pc &         \no \\
\midrule
\end{tabular}
 &
      \begin{tabular}{C{\wid}C{\wid}C{\wid}C{\wid}C{\wid}C{\wid}C{\wid}C{\wid}}
         \no &            \ii &          \pc &            \pc &    \pc &    \pc &           \ii &         \pc \\
         \ii &            \no &          \pc &            \pc &    \pc &    \pc &           \ii &         \pc \\
         \nc &            \nc &          \no &            \ii &    \pc &    \pc &           \nc &         \pc \\
         \nc &            \nc &          \ii &            \no &    \pc &    \pc &           \nc &         \pc \\
         \nc &            \nc &          \nc &            \nc &    \no &    \nc &           \nc &         \pc \\
         \nc &            \nc &          \nc &            \nc &    \pc &    \no &           \nc &         \pc \\
         \ii &            \ii &          \pc &            \pc &    \pc &    \pc &           \no &         \pc \\
         \nc &            \nc &          \nc &            \nc &    \nc &    \nc &           \nc &         \no \\
\midrule
\end{tabular}
\\
      \begin{tabular}{rrC{\wid}C{\wid}C{\wid}C{\wid}C{\wid}C{\wid}C{\wid}C{\wid}}
\multirow{8}{*}{\rotatebox{90}{\textbf{Specificity}}} &
VorEAl (a/c)   &          \no &            \nc &          \nc &            \nc &    \nc &    \nc &           \nc &         \pc \\
& VorEAl (a/c/t) &          \pc &            \no &          \ii &            \nc &    \pc &    \nc &           \ii &         \pc \\
& VorEAl (a/m)   &          \pc &            \ii &          \no &            \nc &    \pc &    \nc &           \ii &         \pc \\
& VorEAl (a/m/t) &          \pc &            \pc &          \pc &            \no &    \pc &    \nc &           \pc &         \pc \\
& NSA-re         &          \pc &            \nc &          \nc &            \nc &    \no &    \nc &           \nc &         \pc \\
& NSA-sp         &          \pc &            \pc &          \pc &            \pc &    \pc &    \no &           \pc &         \pc \\
& one-class SVM  &          \pc &            \ii &          \ii &            \nc &    \pc &    \nc &           \no &         \pc \\
& naïve Bayes    &          \nc &            \nc &          \nc &            \nc &    \nc &    \nc &           \nc &         \no \\
\bottomrule
\end{tabular}
 &
      \begin{tabular}{C{\wid}C{\wid}C{\wid}C{\wid}C{\wid}C{\wid}C{\wid}C{\wid}}
         \no &            \ii &          \nc &            \nc &    \pc &    \pc &           \pc &         \pc \\
         \ii &            \no &          \nc &            \nc &    \pc &    \pc &           \pc &         \pc \\
         \pc &            \pc &          \no &            \nc &    \pc &    \pc &           \pc &         \pc \\
         \pc &            \pc &          \pc &            \no &    \pc &    \pc &           \pc &         \pc \\
         \nc &            \nc &          \nc &            \nc &    \no &    \pc &           \nc &         \nc \\
         \nc &            \nc &          \nc &            \nc &    \nc &    \no &           \nc &         \nc \\
         \nc &            \nc &          \nc &            \nc &    \pc &    \pc &           \no &         \ii \\
         \nc &            \nc &          \nc &            \nc &    \pc &    \pc &           \ii &         \no \\
\bottomrule
\end{tabular}
 &
      \begin{tabular}{C{\wid}C{\wid}C{\wid}C{\wid}C{\wid}C{\wid}C{\wid}C{\wid}}
         \no &            \nc &          \nc &            \nc &    \nc &    \nc &           \nc &         \nc \\
         \pc &            \no &          \nc &            \nc &    \pc &    \nc &           \pc &         \pc \\
         \pc &            \pc &          \no &            \ii &    \pc &    \nc &           \pc &         \pc \\
         \pc &            \pc &          \ii &            \no &    \pc &    \nc &           \pc &         \pc \\
         \pc &            \nc &          \nc &            \nc &    \no &    \nc &           \nc &         \nc \\
         \pc &            \pc &          \pc &            \pc &    \pc &    \no &           \pc &         \pc \\
         \pc &            \nc &          \nc &            \nc &    \pc &    \nc &           \no &         \pc \\
         \pc &            \nc &          \nc &            \nc &    \pc &    \nc &           \nc &         \no \\
\bottomrule
\end{tabular}
 &
      \begin{tabular}{C{\wid}C{\wid}C{\wid}C{\wid}C{\wid}C{\wid}C{\wid}C{\wid}}
         \no &            \ii &          \ii &            \ii &    \pc &    \pc &           \pc &         \pc \\
         \ii &            \no &          \ii &            \ii &    \pc &    \pc &           \pc &         \pc \\
         \ii &            \ii &          \no &            \ii &    \pc &    \pc &           \pc &         \pc \\
         \ii &            \ii &          \ii &            \no &    \pc &    \pc &           \pc &         \pc \\
         \nc &            \nc &          \nc &            \nc &    \no &    \pc &           \pc &         \pc \\
         \nc &            \nc &          \nc &            \nc &    \nc &    \no &           \nc &         \nc \\
         \nc &            \nc &          \nc &            \nc &    \nc &    \pc &           \no &         \pc \\
         \nc &            \nc &          \nc &            \nc &    \nc &    \pc &           \nc &         \no \\
\bottomrule
\end{tabular}
 &
      \begin{tabular}{C{\wid}C{\wid}C{\wid}C{\wid}C{\wid}C{\wid}C{\wid}C{\wid}}
         \no &            \nc &          \nc &            \nc &    \ii &    \nc &           \nc &         \nc \\
         \pc &            \no &          \nc &            \nc &    \pc &    \nc &           \nc &         \nc \\
         \pc &            \pc &          \no &            \nc &    \pc &    \nc &           \nc &         \ii \\
         \pc &            \pc &          \pc &            \no &    \pc &    \ii &           \pc &         \pc \\
         \ii &            \nc &          \nc &            \nc &    \no &    \nc &           \nc &         \nc \\
         \pc &            \pc &          \pc &            \ii &    \pc &    \no &           \pc &         \pc \\
         \pc &            \pc &          \pc &            \nc &    \pc &    \nc &           \no &         \pc \\
         \pc &            \pc &          \ii &            \nc &    \pc &    \nc &           \nc &         \no \\
\bottomrule
\end{tabular}
 &
      \begin{tabular}{C{\wid}C{\wid}C{\wid}C{\wid}C{\wid}C{\wid}C{\wid}C{\wid}}
         \no &            \ii &          \pc &            \pc &    \pc &    \pc &           \ii &         \pc \\
         \ii &            \no &          \pc &            \pc &    \pc &    \ii &           \ii &         \pc \\
         \nc &            \nc &          \no &            \ii &    \pc &    \nc &           \nc &         \pc \\
         \nc &            \nc &          \ii &            \no &    \pc &    \nc &           \nc &         \pc \\
         \nc &            \nc &          \nc &            \nc &    \no &    \nc &           \nc &         \pc \\
         \nc &            \ii &          \pc &            \pc &    \pc &    \no &           \nc &         \pc \\
         \ii &            \ii &          \pc &            \pc &    \pc &    \pc &           \no &         \pc \\
         \nc &            \nc &          \nc &            \nc &    \nc &    \nc &           \nc &         \no \\
\bottomrule
\end{tabular}
\\
    \end{tabular}}
\end{table}

\begin{figure}[tb]
\centering
\begin{minipage}[c]{0.65\columnwidth}
\subfloat[
Summary by metric.\label{fig:sum-metric}
]{
\includegraphics[width=\textwidth]{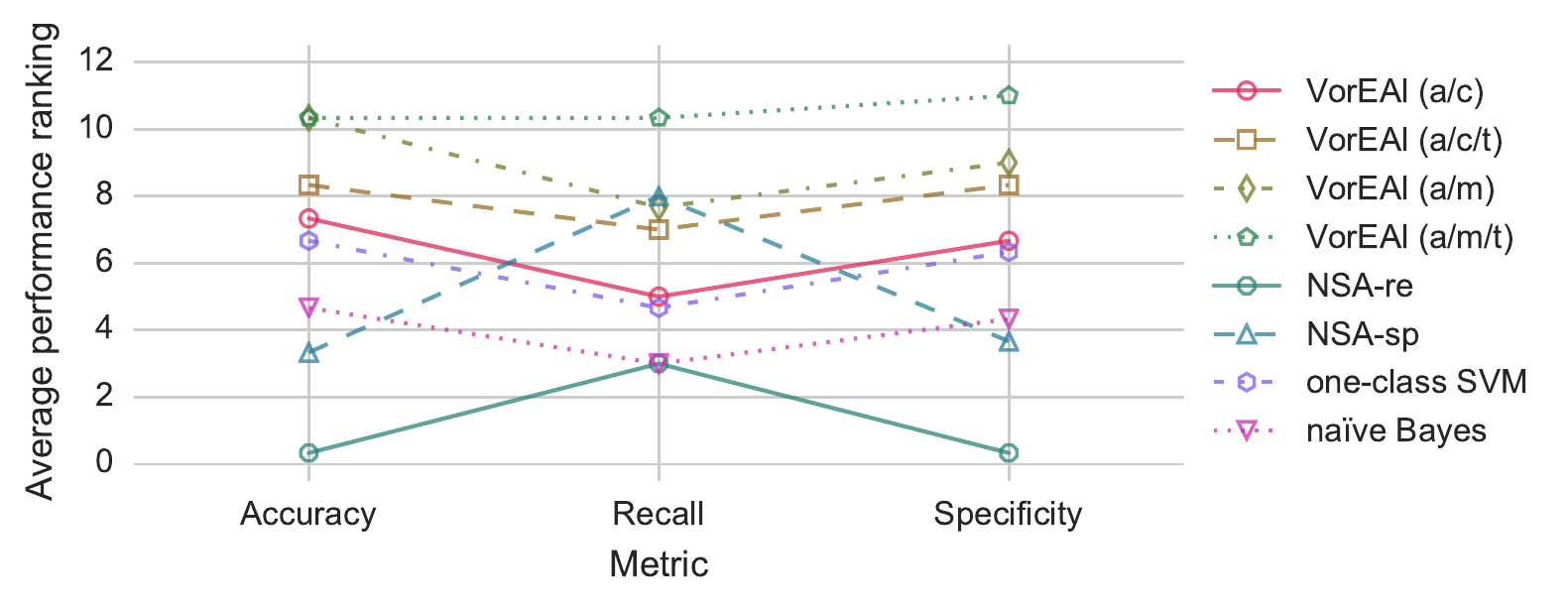}
}\\[-0.65ex]
\subfloat[
Summary by problem.\label{fig:sum-problem}
]{
\includegraphics[width=\textwidth]{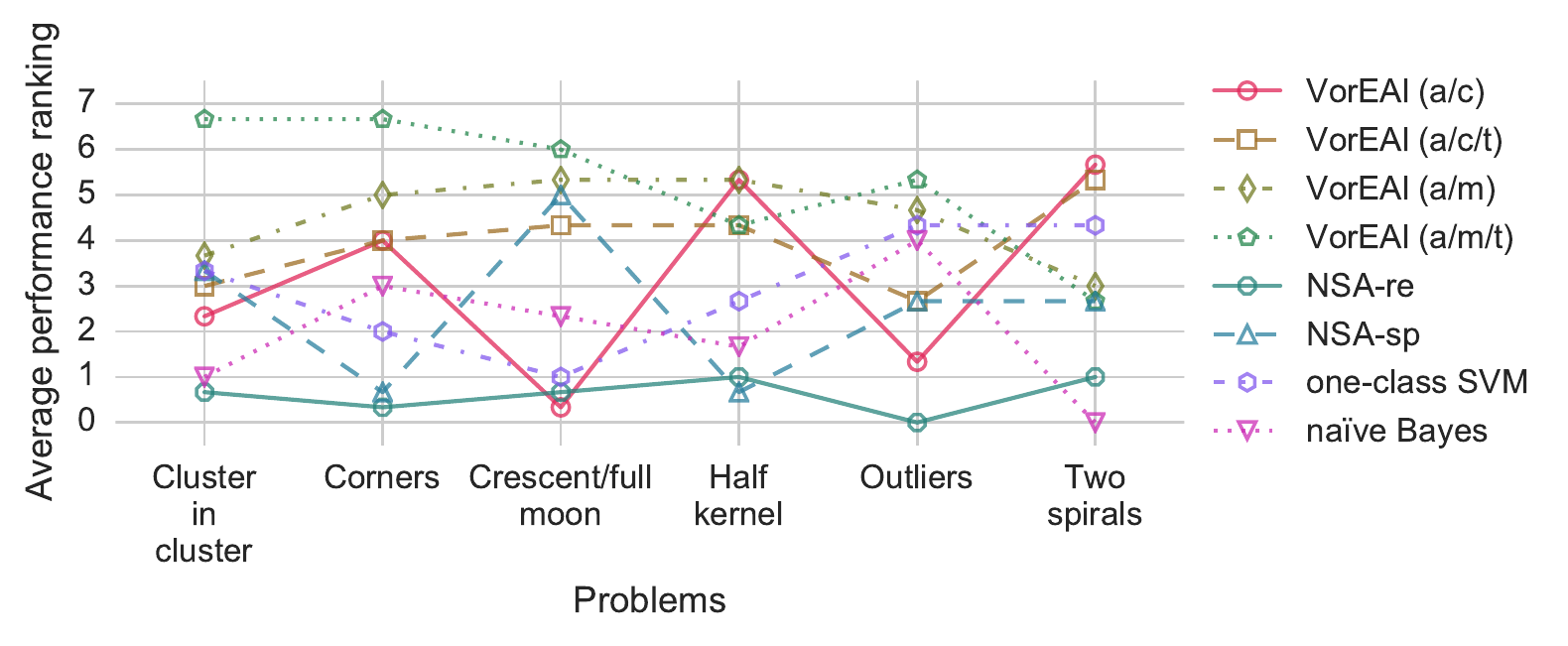}
}
\end{minipage}
\begin{minipage}[c]{0.32\columnwidth}
  \subfloat[Example of an evolved Voronoi diagram for the two spiral problem. Sites that are labeled as anomalies are marked with red triangles ({\color{red}$\blacktriangle$}). Similarly, sites labeled as normal data are indicated with green circles ({\color{green}$\bullet$}). 
  \label{fig:nice-guy}]{\includegraphics[width=\textwidth]{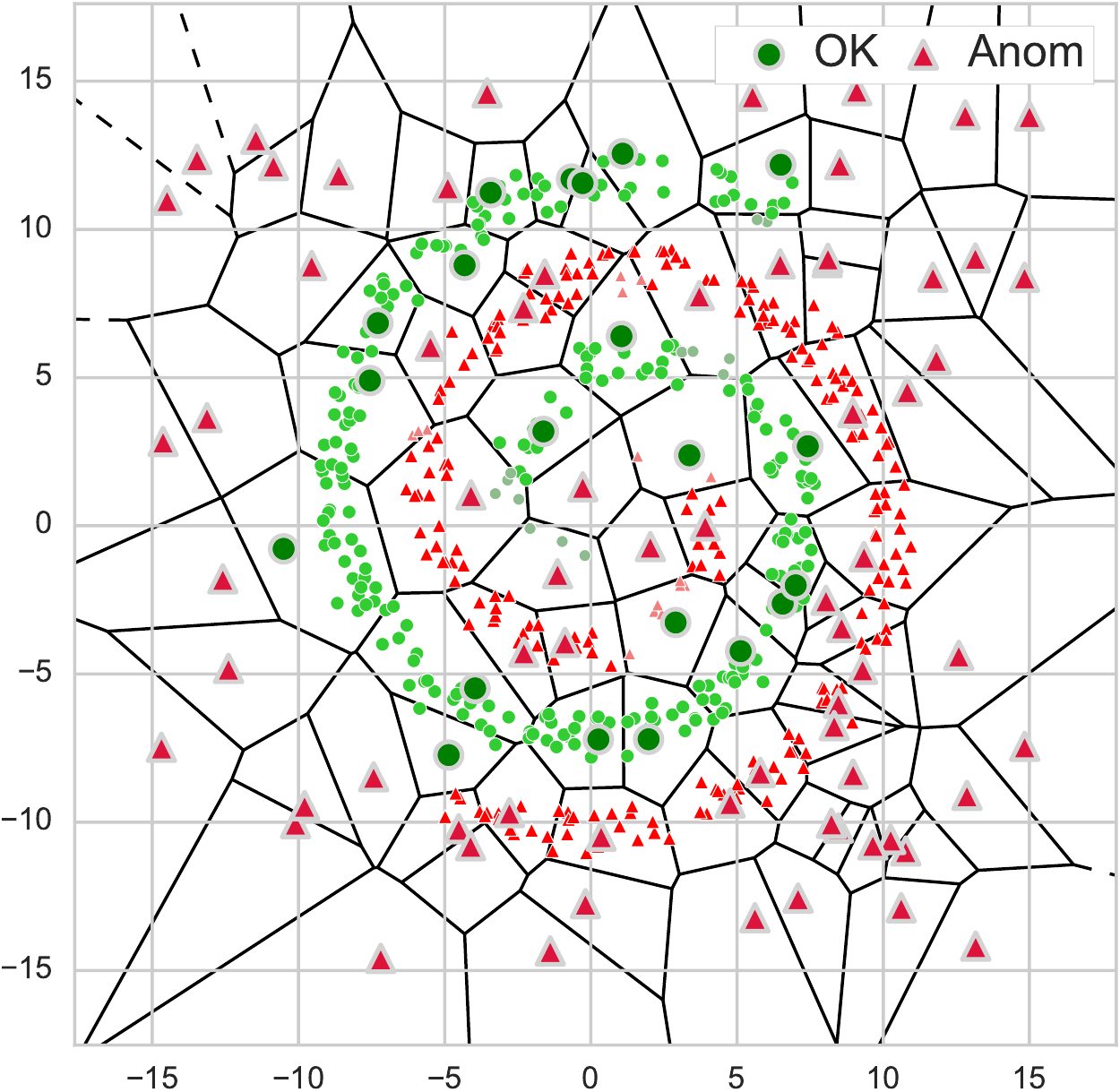}}
\end{minipage}
\caption{Summaries of the statistical tests and an illustrative example.}
\label{fig:problem-sums}
\end{figure}

When many tests are carried out, a comprehensive analysis of the results is rather difficult as it implies cross-examining and comparing the results presented separately. Consequently, we present them in summarized form in Table \ref{tbl:stats}. It should be noted that experiment parameters and results are available online at \url{http://lmarti.com/VorEAl}. To further simplify the understanding of results, why we decided to adopt a more integrative representation like the one proposed in \cite{bader-2010:thesis}.
Figure \ref{fig:problem-sums} summarizes the outcome of the hypothesis tests by grouping them by metric and problem, as explained in the previous section. Here it is clearly visible how VorEAl with multiplicative compactness and total empty volume objectives is generally able to yield better results. Finally, as an illustrative example, we show in Figure \ref{fig:nice-guy} an example of an evolved Voronoi diagram.

\section{Discussion and conclusion}\label{sec:final}

In this paper we have presented VorEAl, a multi-objective evolutionary algorithm that relies on Voronoi diagrams for representation. VorEAl has been devised with the problem of anomaly detection in mind. The experimental results obtained as part of this work point out that this is a promising direction of work. However, there are many areas that should be further studied and explored. From an algorithmic point of view, we should explore other classification objectives (metrics). 

It is important to try other multi-objective fitness assignments, like those based on multi-objective performance indicators or reference points. This last approach is of particular interest as, as we already mentioned, in our case we have an \emph{a priori} known ideal solution that can be used to guide the search. In parallel, work should the done in understanding and reducing the computational complexity of the algorithm. In this direction, we are already working on creating approximative versions of the volume meant to decrease the computational cost of the computation of the objective functions.

\section*{Acknowledgements}

This work has been funded by the project PIA-FSN-P3344-146479. Authors wish to thank the reviewers for their fruitful comments.

\bibliographystyle{splncs03}
\bibliography{biblio,ais,lm,EMOO,pubs}
\end{document}